\documentclass{article}
 
 \pdfoutput=1

  \usepackage{amsmath}
  \usepackage{amssymb}
  \usepackage{color}

  \usepackage{nips14submit_e,times}
  \usepackage{hyperref}
  \usepackage{url}

\title{A unified view of generative models for networks: models, methods, opportunities, and challenges}

\author{
  Abigail Z. Jacobs \\
  Dept of Computer Science\\
  University of Colorado Boulder\\
  \texttt{abigail.jacobs@colorado.edu } \\
  \And
  Aaron Clauset \\
  Dept of Computer Science; BioFrontiers Institute\\
  University of Colorado Boulder\\
  Santa Fe Institute \\
  \texttt{aaron.clauset@colorado.edu} \\
}

\nipsfinalcopy

\begin{document}

\maketitle

\begin{abstract} 
 Research on probabilistic models of networks now spans a wide variety of fields, including physics, sociology, biology, statistics, and machine learning.  These efforts have produced a diverse ecology of models and methods. Despite this diversity, many of these models share a common underlying structure:\ pairwise interactions (edges) are generated with probability conditional on latent vertex attributes. Differences between models generally stem from different philosophical choices about how to learn from data or different empirically-motivated goals. The highly interdisciplinary nature of work on these generative models, however, has inhibited the development of a unified view of their similarities and differences. For instance, novel theoretical models and optimization techniques developed in machine learning are largely unknown within the social and biological sciences, which have instead emphasized model interpretability. Here, we describe a unified view of generative models for networks that draws together many of these disparate threads and highlights the fundamental similarities and differences that span these fields. We then describe a number of opportunities and challenges for future work that are revealed by this view.
\end{abstract}

\section{Introduction}

The quantitative study of the structure, dynamics, and function of real-world networks is an emerging interdisciplinary field that spans nearly every domain of science. Generative models, in which we define a structured probability distribution over all graphs, are a powerful and increasingly popular approach for studying real-world networks. Work on generative models now spans machine learning, statistics, the social sciences, ecology, and statistical physics. However, as a result of their different traditions and goals, these communities have developed a range of different generative models, including latent position models~\cite{Hoff2002}, block models~\cite{Airoldi2008a}, and feature models~\cite{Miller2009}. Furthermore, these communities use different methods to learn such models from data, including frequentist, Bayesian, and nonparametric methods. Despite this large and diverse ecology of approaches, formulating a coherent understanding of whether and how these models are related remains an open challenge (notwithstanding several thoughtful efforts~\cite{Hoff2007,Thomas2009,Schmidt2013a,Goldenberg:2010:SSN:1734794.1734795}). We focus on the general framework of modeling conditionally independent edges between vertices 
 with latent vertex attributes.

Here, we survey and organize this diversity of models and methods, and then introduce a unified view of the field. This view positions the three main classes of generative models for networks as special cases of a single coherent framework, positions the three major approaches to learning these models as different philosophical choices, and yields new insights from translation, such as theoretical properties, interpretation, and new inference algorithms. We close with a discussion of the challenges and opportunities this view presents for the field.

\section{A walking tour of existing models}\label{sec:litreview} 

The following brief tour of generative models for networks illustrates the large diversity of approaches, as well as the different traditions used across disciplines. Despite different points of origin and assumptions for many of these models, we emphasize their underlying similarity throughout this tour of models. We note that dynamic networks, multiplex networks, networks with metadata, or networks produced from physical processes~\cite{newman:networks:book} can often be modeled in a straightforward manner using the model classes described below.

The first class of models is the latent space models. The canonical latent space model was introduced in statistics and mathematical sociology by Hoff and colleagues~\cite{Hoff2002}. In this model, vertices take latent positions $z_i \in \mathbb{R}^K$ for a $K$-dimensional space and edges are generated with a probability that depends only on the Euclidean distance between vertex positions. Subsequently, Hoff  introduced a more flexible construction called ``the eigenmodel'' that captures both the latent space model and latent block models~\cite{Hoff2007}. In general, latent space models assume latent continuous vertex attributes with edge probabilities given by a distance function on those attributes. Moreover, by relaxing the continuous (or Euclidean~\cite{krioukov2010hyperbolic}) space assumption and allowing edge probabilities to be determined by a generic function of those attributes, we can recover as special cases each of the two model classes discussed below. Thus, latent space models effectively subsume all generative models for networks.

In the canonical latent space model, connection probabilities are assumed to decrease with latent distance, which induces strong assortativity in the resulting networks (\textit{like} vertices connect with \textit{like}). This pattern implies that vertices can be clustered by their latent positions. It then is natural to define a hierarchical model that explicitly represents such clusters within the latent space model~\cite{Handcock2007}. But, homophily can be a misleading assumption for many real networks, which sometimes exhibit ordered (status-oriented) or disassortative patterns (\textit{dislike} connects with \textit{dislike}). Disassortativity in particular is common in many biological, technological, and economic networks. For example, the probabilistic niche model from ecology is a general latent space model that can naturally produce assortative, ordered, or disassortative latent clustering by dissociating sending and receiving latent positions of each vertex~\cite{Williams2011}. 

Finally, exponential random graph models in the sociological literature give network likelihood an exponential family form, typically as a function of network statistics; whenever they employ latent variables, they also fall within the latent space model class~\cite{Robins2007,Schmidt2013a}.

Our second class is the block models, which use a latent space defined by categorical variables rather than continuous variables:\ vertices take positions $z_i$ in some latent space $\{0,1\}^K$ for $K$ blocks, where $||z_i||_1 = 1$. These models are often used to cluster vertices into roughly homogeneous groups or classes, a task akin to ``community detection'' in network science~\cite{Karrer2011,newman:networks:book}. Block models are by far the most well-explored and broadly used class of generative models for networks.

Block models originated in mathematical sociology~\cite{Holland1983,Wang1987,Nowicki2001} and represent active, albeit somewhat disconnected, areas of research in sociology, machine learning, and statistical physics~\cite{Schmidt2013a,Goldenberg:2010:SSN:1734794.1734795,newman:networks:book}. In a block model, vertices can belong to either exactly one latent block (hard clustering) or have partial memberships in multiple blocks (overlapping or mixed membership). Vertices within a block are stochastically equivalent, meaning any pair of vertices in a given block have the same probabilities to connect to all other vertices in the network, and block interactions describe a coarse-graining of the network interactions. If vertices can have mixed membership, vertices instead take latent positions in the simplex~\cite{Airoldi2008a}. In nonparametric variations, $K$ is first sampled, otherwise fixed in advance~\cite{Schmidt2013a,Kemp2006}. The simplest block model is the Erd\H{o}s-R\'enyi random graph, in which $K=1$. Larger models can exhibit a rich variety of block-level patterns, including assortative, disassortative, core-periphery, ordered, bipartite structure~\cite{Larremore2014} and overlapping groups~\cite{Airoldi2008a}. Furthermore, block models can be extended to directly model other structural information, including degree heterogeneity~\cite{Karrer2011}, edge-weights~\cite{Aicher2014}, social status~\cite{Ball2012}, or a growing number of blocks~\cite{Schmidt2013a,Kemp2006,Choi2012}. It is worthwhile to note that block models include topic models as a special case~\cite{Airoldi2008a}.

The third class is the latent feature models. This class is a natural progression from latent blocks and explicitly highlights the connection to the canonical latent space model. Instead of dividing total membership into one or across a mixture of groups, latent feature models allow vertices to have arbitrarily many unique features, typically binary, so that vertex attributes have the form $z_i \in \{0,1\}^K$, where $||z_i||_1 \le K$. Machine learning offers a rich literature on latent feature models that extends naturally to relational data, i.e., networks~\cite{Miller2009}. Edge probabilities are given as a function of a weighted sum of feature vectors: by allowing these weights to be positive or negative, latent feature models can allow vertex features to combine assortatively or disassortatively. Richer structure, including hierarchies on the features, can also be imposed~\cite{Schmidt2013a,Palla2012}. Alternatively, fixed-$K$ (parametric) variations combine features with block models~\cite{Kim2012d} or in a regression framework~\cite{Krivitsky2009a}.

A fourth class includes models defined on a latent hierarchical or tree-like space. Although such a latent structure may appear distinct from the previous model classes, these models of hierarchical structure are in fact a special subset of the other classes. For instance, we can directly impose a hierarchical organization on the latent block or feature models. Some models can be explicitly defined using an underlying tree structure, in which vertices in the network are leaves and common ancestors determine the probability of connection, as in the (nonparametric) hierarchical random graph model~\cite{Clauset2008} and the Bayesian and nonparametric variations~\cite{Roy2007,Roy2009,Schmidt2013a}. Each of these models uses a piece-wise constant underlying space~\cite{Orbanz2014}, or a discrete latent space model, in which ultrametric distances between vertices determine the probability of connection~\cite{Snijders2011}; the continuous version can be captured by hyperbolic geometry of a graph~\cite{krioukov2010hyperbolic,asta2014geometric}.

\section{Philosophies, representations, and unifications}

In order to learn a particular generative model using real network data, we must also choose a learning paradigm that defines the relationship between an observed network and the various parameters of the model (including complexity parameters like $K$). Here, we characterize the three major approaches for learning network models in terms of their different philosophical assumptions, and illustrate how moving a model between paradigms can yield insights into theory and interpretation, and shed new light on both practical and efficient application of these models to network data.

Generative models for networks can be defined as parametric, frequentist or Bayesian, or under the Bayesian nonparametric paradigms. (Frequentist nonparametric methods are available as well, although relatively uncommon in this context. There is a bit of this flavor in some recent papers~\cite{Ball2012,asta2014geometric,Chatterjee2012,Olhede2014a}, however.) However, different communities usually favor one paradigm over the others, which generally reflects differences in traditions and goals as much as it reflects any objective rationale. As popular as hierarchical Bayesian models are in machine learning, frequentist perspectives still dominate the ecology, economics and statistical physics literatures, often with an emphasis on interpretability. Some network models have appeared in multiple communities, and these can be independent rediscoveries under differing paradigms or are intentional adaptations from one paradigm to another, although this is not always obvious. There is thus much to be gained by unifying various network models within a single paradigm of learning from data.

Past efforts at unification have generally focused on placing large classes of models within a common representation and restricted set of philosophical choices. For instance, the eigenmodel~\cite{Hoff2007} captures both latent space and block models, which can also be represented within a generalized linear model schema~\cite{Thomas2009}. Another general approach is matrix estimation~\cite{Chatterjee2012}, which can be applied broadly to networks in their adjacency matrix representation, and is in the tradition of nonparametric graph limit (graphon) estimation~\cite{Yang2014a,Bickel2011,Wolfe2013,Bickel2009,Lloyd-randpriors-2012}.

Bayesian nonparametric models are also increasingly popular, particularly in the machine learning community, and many of the models previously described can be unified under this framework~\cite{Schmidt2013a,Orbanz2014,Lloyd-randpriors-2012}. Placing generative models for networks under the Bayesian nonparametric umbrella allows the model implies a particular method of model selection (which is effectively subsumed within the inference step). Bayesian nonparametric models allow the structural parameters to change, sometimes dramatically so, as more data is observed, e.g., with an increasing number of blocks or features. In practice, this means we begin with an infinite dimensional space and move to a finite representation. Practical application also requires a choice of priors that are tractable---though the consequences of which remain unclear, e.g., for consistency and practical performance. On the other hand, efforts under the Bayesian nonparametric framework have successfully led to novel machinery for modeling, inference, and theory~\cite{Caron2014,Orbanz2014,Bickel2009}.

To illustrate the more general theoretical perspective to be gained from this broad view of learning network models, we explore recent work defining the underlying theoretical basis for generative models for networks via the graphon. This approach is not without its limitations, however, and we then explore a related but distinct approach based on continuous processes.

\subsection{Unification under the graphon}\label{subsec:graphons}

A reasonable and desirable property of generative models for networks is that they should not depend on the order in which we observe data, i.e., that our models are \textit{exchangeable}. Network data, presented as a (random) adjacency matrix, requires joint exchangeability, so row and column identities in the matrix are jointly preserved under permutations of the data.\footnote{Bipartite networks and feature data require separable exchangeability, as row and column identities need not be related. Joint exchangeability is a special case that notably requires symmetry of the graphon.} Exchangeability as a requirement leads to representation theorems. For exchangeable \textit{sequences}, de Finetti's theorem implies that they can be represented by an underlying  i.i.d.\ mixture of random variables. For a jointly exchangeable adjacency matrix, the Aldous-Hoover theorem implies the existence of a random measurable function, given in the following manner: for each vertex $i$, draw $U_i$ uniformly at random from the unit interval $[0,1]$, and for each pair of vertices $i,j$, draw $U_{ij}$ uniformly at random from $[0,1]$. There then exists a function $\omega:[0,1]^2 \rightarrow [0,1]$ such that the adjacency matrix entries follow 
$
X_{ij} \,{\buildrel d \over =}\, \mathbf{1}_{U_{ij} < \omega(U_i,U_j)}
$. 
We call this function $\omega$ the \textit{graphon}~\cite{Orbanz2014}. Equivalently, the graphon is the limit object of a sequence of graphs~\cite{Diaconis2007,Austin2008,Lovasz2012}, which can then in theory be directly estimated nonparametrically~\cite{Yang2014a,Wolfe2013,Bickel2009,Bickel2011,Lloyd-randpriors-2012}.

From this perspective, exchangeability for network data implies the existence of a latent variable generative model, and furthermore that in such a model, edges are conditionally independent. The latent variable models discussed above can be justified under this framework~\cite{Hoff2007}.

However, an important consequence of the graphon construction is that graphs will be almost surely dense ($\Theta(n^2)$ edges) or trivially empty~\cite{Orbanz2014,Lovasz2012}. That is, exchangeability of a network model implies that we are in the regime of dense networks. This is problematic because most real-world networks, in fact, are sparse (vertices have constant or very slowing-growing degree; graphs have $O(n)$ edges). Most of the models discussed in our walking tour fall within the jointly exchangeable framework of Aldous-Hoover. This would seem to imply that all generative models for networks are misspecified, despite their many successful practical applications. One solution to this fundamental problem is to abandon exchangeability in favor of alternative properties~\cite{Orbanz2014}, or to attempt to escape the Aldous-Hoover representation entirely~\cite{Caron2014}.

\subsection{Unification under continuous space}\label{subsec:caronfox}

To move beyond the graphon, Caron and Fox lift the representation of discrete adjacency matrices into a continuous space~\cite{Caron2014}. Under this formulation, graphs are generated purely as a point process. By representing a graph as a sample from a continuous-time process, the Caron-Fox formulation sidesteps the dense-graph implication of the Aldous-Hoover approach while still preserving joint exchangeability. Instead of a random measurable function on the unit square, this approach implies, 
due to Kallenberg, a representation as a mixture of random functions.

This alternative approach is promising, as it presents the possibility that generative models for networks are not all misspecified, so long as they can be reformulated as a point process model for networks. To demonstrate this approach, Caron and Fox 
describe several choices of parametrization that correspond to an Erd\H{o}s-R\'enyi $G(n,p)$ model, graphon model, and the configuration model for random graphs with specified degree sequence. Each vertex has a latent variable weight parameter corresponding to sociability, which is analogous to degree ``propensity'' in the popular degree-corrected stochastic block model~\cite{Karrer2011}. This model parameterization connects their model explicitly to the family of physical models with specified degree distribution~\cite{newman:networks:book}.

Introducing higher order structure, such as community, ordered, or hierarchical structure, has not yet been accomplished, and will be a natural and productive extension of these models. Hierarchical models that are based on the point process model seem likely to yield new theoretical insights and inference algorithms. On the other hand, more is currently known about the graphon and the pathologies of discrete graph models than about the continuous-space approach. Thus, there may be as-yet-unknown difficulties lurking within the point process model. One immediate issue is that this model only yields graphs with superlinear density $O(n^\alpha),\ 1<\alpha < 2$, that is, it does not model extremely sparse graphs ($O(n)$ edges for $O(n)$ vertices)~\cite{Caron2014}. It will be crucial to assess the practical implications of these models to understand if, when, and how they break down on real-world network data.

\section{Challenges and opportunities}

Moving models between learning paradigms has already been a productive source of new ways to model networks. However, without a broader interest in the application of these models or in the underlying theoretical questions about networks, such model development risks being primarily a fluency exercise in philosophical bases and notation. The broad view of generative models for networks described above reveals a number of challenges and opportunities for the field, many of which are directly related to moving between representations and translating insights across fields.

\paragraph{Interpretability.} The interpretability of model parameters, model structure, and their relationship to network data are crucial for inference about real-world complex systems. Identifying interpretable underlying structure is a strong motivation for many of the applications of network models in the sciences. For instance, the structure of social networks is generally believed to be driven by latent spaces, including geography~\cite{Liben-Nowell2005}, socioeconomic status~\cite{Eagle2010}, and popularity~\cite{Ball2012}. And, models of biological networks generally seek structural ``modules'' that are meaningfully related to biological function, e.g., species with similar feeding patterns in food webs, the different disease phenotypes of the malaria parasite, or proteins with similar cellular functions~\cite{Clauset2008,Larremore2014,Airoldi2008a}.

Models motivated by specific applications can be useful for understanding how to interpret the different patterns of large-scale network organization encoded by different models~\cite{Goldenberg:2010:SSN:1734794.1734795}. Extremely general models often have limited interpretability, and thus also limited utility for scientific applications. That is, we tend to trade-off between model interpretability and model generality (not necessarily model complexity). For instance, approximating the graphon with Gaussian processes~\cite{Lloyd-randpriors-2012} is quite general but such a model cannot easily produce useful insights about the underlying mechanisms that generated any particular network. Thus, an opportunity and a challenge is the adaptation of structural metaphors from applied domains to general models, allowing us to leverage the corresponding interpretations, e.g., social group structure to block models or ecological niche structure to latent space models~\cite{Goldenberg:2010:SSN:1734794.1734795,Williams2011}.

\paragraph{Homophily.} \textit{Assortativity} is the tendency in observational data that vertices that are connected have similar attributes. Many generative models for networks assume that similarity between vertices is what generates edges (a process called \textit{homophily}), and conversely, that we can infer vertex similarity directly from the structure of a network. That is, these models assume homophily is the underlying generative mechanism. However, assortativity also occurs when vertices become more similar as a result of an existing connection (a process sometimes called ``social contagion,'' although that name has unfortunate connotations). In particular, influence can also produce assortativity, and few generative models allow for both mechanisms. Deciding which of the two governs a particular system is sometimes scientifically attractive but it is also generally statistically impossible~\cite{Shalizi2011a}.

Observed or latent attributes can drive assortativity directly, as in homophily, but homophily and structural similarity are easily confounded as well. (There remains another idea to conflate:\ detecting assortativity is not the same as predicting, given some vertex features, that similar vertices will be connected. Assortativity only guarantees that given an edge, vertex features are likely to be similar.) Furthermore, vertex attributes and network structure may be unrelated. Fosdick and Hoff~\cite{Fosdick2013} designed a framework to first test for dependence between vertex attributes and network structure, and then model network structure and attributes jointly. Even then, dependence between attributes and latent network structure is not causality. Latent positions correlated with known metadata may be correlated with the true causal mechanism, e.g., jointly caused by a true but unknown mechanism~\cite{VonLuxberg2012}. Although latent space models do not solve the correlation-causation problem, they can be a useful way to instrumentalize these processes.

In the opposite direction, we can leverage patterns of assortativity to make useful predictions. Recommender systems are built on this notion, where we can infer profiles of interests given a user profile. In sociology for instance, McCormick and Zheng use the latent space model~\cite{Hoff2002} to model the distribution of latent social positions of partially unobserved populations~\cite{Mccormick2013-ARD}. They use these distributions to infer demographic profiles of underrepresented populations, i.e., in a setting where such assumptions about generalizability and assortativity may be justified.

\paragraph{Model selection} Practical model selection for models of network data remains an open challenge. This includes finding the dimensionality of the latent space or number of blocks or features: different choices can easily introduce a linear number of new parameters. A number of methods have been applied, including AIC~\cite{Williams2011} and BIC~\cite{Airoldi2008a} (both of which are known to fail in our context~\cite{Yan2014}); MDL~\cite{Peixoto2013,peixoto2014model}; Bayes factors~\cite{Aicher2014,Hofman2008,Handcock2007}; and likelihood ratios~\cite{Yan2014}. Bayesian nonparametric models move the modeling selection task \textit{inside} the model, where dimensionality becomes a parameter to be inferred. In general, it is unclear when these methods fail. Potentially inappropriate assumptions made during inference (e.g., assortativity of latent blocks~\cite{Gopalan2012}) and lack of effective validation~\cite{VonLuxberg2012} complicate this already non-trivial challenge.

\paragraph{Model checking} Even if our favorite model selection technique succeeds, how do we know if the latent variable representation we have inferred is reasonable? One answer is metadata: for example, if the recovered latent space corresponds to vertex information that was excluded from the model~\cite{Fosdick2013}. On the other hand, using unsupervised methods to find patterns that correlate with `ground truth' can be problematic, depending on how we validate and interpret the inferred vertex distribution and latent space~\cite{VonLuxberg2012}.

Simulation-based model checking can be used to further probe the reasonableness of our models, that is, th egoodness of fit~\cite{Hunter2008,Gelman2013}. Model checking can then expand beyond link prediction: e.g., by measuring the structural similarity of resampled graphs to the original data. This manner of model checking is also reminiscent of approximate Bayesian computation (ABC), which avoids intractable optimization problems by comparing other properties from the resampled model. Model checking in this manner can be used further to compare approximate inference methods. Finally, model checking provides a robust critical method to investigate hypotheses~\cite{Gelman2013}.

Model checking also points the way to potential method to compare networks, which is currently an open challenge for working with network data. However, we currently lack a sensible general framework with which to compare networks. For networks suspected to have come from the same generative model, one can resample networks and compare the networks of interest. (Even this is dependent on the model specification, however.  Exponential random graph models, for example, lack consistency under subsampling~\cite{Shalizi2013a}.) If the graphon is used directly, one would be able to directly compare networks of different sizes, but this requires estimating the graphon directly such that samples could be compared. To sidestep this issue, Asta and Shalizi recently proposed a nonparametric graph comparison method based on hyperbolic latent space models~\cite{asta2014geometric}.

\paragraph{Optimization} Models for networks suffer dependencies among data and model parameters, and tend to scale poorly or suffer from strong degeneracies. MCMC methods remain popular across domains, despite poor scalability~\cite{Goldenberg:2010:SSN:1734794.1734795,Robins2007,Williams2011,newman:networks:book}. Variational methods and belief propagation (message passing) have become popular in both statistical physics and machine learning (e.g.,~\cite{Airoldi2008a,Salter-Townshend2013,Aicher2014,newman:networks:book}), including recent extension to stochastic variational inference~\cite{Gopalan2012}. Variational methods are much faster, but the costs on these models of their mean-field approximation (independence assumptions) are not well understood. In addition, a number of algorithms make assumptions that cannot be widely true, such as strictly assortative clustering~\cite{Gopalan2012}. Belief propagation, likewise, is very efficient but necessarily approximate and only suitable for (tree-like) sparse graphs. Pseudolikelihood methods have been developed for latent feature models~\cite{Reed2013}, block models~\cite{Amini2013}, and the latent space model~\cite{Raftery2012}. Currently, Bayesian nonparametric models tend to suffer during optimization and generally scale poorly, but we expect this to be an active area moving forward~\cite{Schmidt2013a}.

\paragraph{Model use and accessibility} Making new models relevant to the wider communities using network analysis requires making them easily accessible in language, construction, and implementation. One challenge already troubling inference and analysis for these models comes from the inherent symmetries in the model construction. Translational and reflectional symmetries, as well as other forms of non-identifiability, are a practical issue for comparing and combining point estimates and posterior probabilities (i.e., model selection and comparison). These symmetries contribute to the ruggedness of the likelihood space, which further challenges the validity of variational assumptions for model inference~\cite{Nishihara2013}. This---and the challenges already discussed---will necessarily create challenges for automatic, rapid model construction and evaluation, as in probabilistic programming~\cite{Nishihara2013}. Sometimes this can be resolved simply with a canonical (but arbitrary) setting (e.g.,~\cite{Clauset2008}) but even this can be theoretically non-trivial~\cite{Yang2014a}. Even if model-fitting tools become widely available, applied network analysis still generally lacks best practices for coping with model non-identifiabilities. To meaningfully impact change, such developments will need to be developed jointly between theoreticians and practitioners.

\section{Discussion}

Generative models are a powerful and increasingly popular approach for understanding the structure of networks. They are also studied or used by researchers spanning a surprisingly diverse set of fields, including both method-oriented fields like machine learning and question-oriented fields like ecology and physics. This diversity has produced a broad variety of models and a highly fragmented literature. It has also slowed the development of a coherent understanding of the theoretical basis and practical applications of these models. Most of the specific generative models developed across these literatures, however, can be viewed as being special cases of a general latent space model, where the latent space may have certain restrictions or characteristics, and employing one of three particular learning paradigms for working with actual data. This view is not a grand unified mathematical theory, but it does provide a coherent understanding of how different models in different fields are related. This view also highlights a number of interesting challenges and opportunities for future work on generative models for networks.

Even after connecting these perspectives, we still face the challenge of evaluating the theoretical and practical costs and benefits of different sets of modeling assumptions or choices over one another. For example, how can we understand what is being traded off when we use a latent block model under a Bayesian nonparametric learning paradigm versus a general latent space model under a frequentist learning paradigm? Even when pursuing the same basic goals, different communities of researchers and even different individuals can rationalize different modeling choices, e.g., which priors to use for regularization, how to manage model parameter uncertainty, or when to employ frequentist-style hypothesis testing.

At their core, many of these choices reflect fundamental beliefs about the nature of data, e.g., the choice of distance function or the choice of representation for class or feature memberships. Other choices reflect fundamental beliefs about the nature of models, such as believing that the number of parameters should grow with larger sample sizes or that domain knowledge should be incorporated into the expected distributions and relationships between variables. The costs of these choices are often unknown in part because the modeling goals are usually unstated, but the consequences can be quite real. For instance, what is the empirical cost of choosing mathematically tractable priors for Bayesian nonparametric models? How much more or less robust is low-level parametric modeling versus a general nonparametric approximation?

One of the genuine advantages of generative models for networks is that they explicitly encode our beliefs about both the nature of the data and the nature of our models, which allows us to carefully quantify parameter and model uncertainty and to interpret the results with respect to the data generating process. What we lack, however, is an effective, principled way to evaluate the costs and benefits of specific choices with respect to specific goals.

Making progress on this fundamental issue would shed considerable light on all aspects of generative models for networks. Connecting models to high-level representations, for discrete random graphs~\cite{Orbanz2014,Goldenberg:2010:SSN:1734794.1734795} and continuous space representations~\cite{Caron2014} provides an illustrative example of exploring these boundaries. Translating modeling techniques such as block structure and hierarchical structure to the point process model should provide an exciting new ground for model development, borrowing interpretation and developments already established on simpler, discrete spaces and physical, ecological, and social processes. Finally, jointly building theoretical foundations for these models while representing more complex structure in network data will help unite and benefit network science, both in applied and theoretical domains.

\subsubsection*{Acknowledgments}
This work was supported by the US AFOSR and DARPA grant number FA9550-12-1-0432 (AZJ, AC) and the NSF Graduate Research Fellowship award number DGE 1144083 (AZJ).


{\small


\begin{thebibliography}{10}

\bibitem{Hoff2002}
P.~D. Hoff, A.~E. Raftery, and M.~S. Handcock, ``{Latent space approaches to
  social network analysis},'' {\em Journal of the American Statistical
  Association}, vol.~97, no.~460, pp.~1090--1098, 2002.

\bibitem{Airoldi2008a}
E.~M. Airoldi, D.~M. Blei, S.~E. Fienberg, and E.~P. Xing, ``{Mixed Membership
  Stochastic Blockmodels},'' {\em Journal of Machine Learning Research},
  vol.~9, pp.~1981--2014, 2008.

\bibitem{Miller2009}
K.~Miller, T.~L. Griffiths, and M.~I. Jordan, ``{Nonparametric latent feature
  models for link prediction},'' {\em NIPS}, 2009.

\bibitem{Hoff2007}
P.~D. Hoff, ``{Modeling homophily and stochastic equivalence in symmetric
  relational data},'' {\em NIPS}, 2007.

\bibitem{Thomas2009}
A.~C. Thomas, {\em {Hierarchical Models for Relational Data}}.
\newblock PhD thesis, Harvard University, 2009.

\bibitem{Schmidt2013a}
M.~N. Schmidt and M.~M{\o}rup, ``{Nonparametric Bayesian Modeling of Complex
  Networks},'' {\em IEEE Signal Processing Magazine}, pp.~110--128, 2013.

\bibitem{Goldenberg:2010:SSN:1734794.1734795}
A.~Goldenberg, A.~X. Zheng, S.~E. Fienberg, and E.~M. Airoldi, ``{A Survey of
  Statistical Network Models},'' {\em Found. Trends Mach. Learn.}, vol.~2,
  no.~2, pp.~129--233, 2010.

\bibitem{newman:networks:book}
M.~E.~J. Newman, {\em {Networks: An Introduction}}.
\newblock Oxford, UK: Oxford University Press, 2010.

\bibitem{krioukov2010hyperbolic}
D.~Krioukov, F.~Papadopoulos, M.~Kitsak, A.~Vahdat, and M.~Bogu\~n\'a,
  ``Hyperbolic geometry of complex networks,'' {\em Phys. Rev. E}, vol.~82,
  p.~036106, 2010.

\bibitem{Handcock2007}
M.~S. Handcock, A.~E. Raftery, and J.~M. Tantrum, ``{Model-based clustering for
  social networks},'' {\em Journal of the Royal Statistical Society: Series A},
  vol.~170, no.~2, pp.~301--354, 2007.

\bibitem{Williams2011}
R.~J. Williams and D.~W. Purves, ``{The probabilistic niche model reveals
  substantial variation in the niche structure of empirical food webs.},'' {\em
  Ecology}, vol.~92, no.~9, pp.~1849--57, 2011.

\bibitem{Robins2007}
G.~Robins, P.~Pattison, Y.~Kalish, and D.~Lusher, ``{An introduction to
  exponential random graph (p*) models for social networks},'' {\em Social
  Networks}, vol.~29, no.~2, pp.~173--191, 2007.

\bibitem{Karrer2011}
B.~Karrer and M.~E.~J. Newman, ``{Stochastic blockmodels and community
  structure in networks},'' {\em Physical Review E}, vol.~83, no.~1, p.~016017,
  2011.

\bibitem{Holland1983}
P.~W. Holland, K.~B. Laskey, and S.~Leinhardt, ``{Stochastic blockmodels: First
  steps},'' {\em Social Networks}, vol.~5, no.~2, pp.~109--137, 1983.

\bibitem{Wang1987}
Y.~J. Wang and G.~Y. Wong, ``{Stochastic Blockmodels for Directed Graphs},''
  {\em Journal of the American Statistical Association}, vol.~82, no.~397,
  pp.~8--19, 1987.

\bibitem{Nowicki2001}
K.~Nowicki and T.~A.~B. Snijders, ``{Estimation and Prediction for Stochastic
  Blockstructures},'' {\em Journal of the American Statistical Association},
  vol.~96, no.~455, pp.~1077--1087, 2001.

\bibitem{Kemp2006}
C.~Kemp, J.~B. Tenenbaum, T.~L. Griffiths, T.~Yamada, and N.~Ueda, ``{Learning
  systems of concepts with an infinite relational model},'' {\em AAAI}, 2006.

\bibitem{Larremore2014}
D.~B. Larremore, A.~Clauset, and A.~Z. Jacobs, ``{Efficiently inferring
  community structure in bipartite networks},'' {\em Physical Review E},
  vol.~90, no.~1, p.~012805, 2014.

\bibitem{Aicher2014}
C.~Aicher, A.~Z. Jacobs, and A.~Clauset, ``{Learning Latent Block Structure in
  Weighted Networks},'' {\em Journal of Complex Networks, \textnormal{to
  appear. }arxiv:1404.0431}, 2014.

\bibitem{Ball2012}
B.~Ball and M.~E.~J. Newman, ``{Friendship networks and social status},'' {\em
  Network Science}, vol.~1, no.~01, pp.~16--30, 2013.

\bibitem{Choi2012}
D.~S. Choi, P.~J. Wolfe, and E.~M. Airoldi, ``{Stochastic blockmodels with a
  growing number of classes},'' {\em Biometrika}, pp.~273--284,
  2012.

\bibitem{Palla2012}
K.~Palla, D.~Knowles, and Z.~Ghahramani, ``{An infinite latent attribute model
  for network data},'' {\em ICML}, 2012.

\bibitem{Kim2012d}
M.~Kim and J.~Leskovec, ``{Multiplicative Attribute Graph Model of Real-World
  Networks},'' {\em Internet Mathematics}, vol.~8, no.~1-2, pp.~113--160, 2012.

\bibitem{Krivitsky2009a}
P.~D. Hoff, ``{Multiplicative latent factor models for description and
  prediction of social networks},'' {\em Computational \& Mathematical
  Organization Theory}, vol.~15, no.~4, pp.~261--272, 2009.

\bibitem{Clauset2008}
A.~Clauset, C.~Moore, and M.~E.~J. Newman, ``{Hierarchical structure and the
  prediction of missing links in networks},'' {\em Nature}, vol.~453, no.~7181,
  pp.~98--101, 2008.

\bibitem{Roy2007}
D.~M. Roy, C.~Kemp, V.~K. Mansinghka, and J.~B. Tenenbaum, ``{Learning
  annotated hierarchies from relational data},'' {\em NIPS}, 2007.

\bibitem{Roy2009}
D.~M. Roy and Y.~W. Teh, ``{The Mondrian Process},'' {\em NIPS}, 2009.

\bibitem{Orbanz2014}
P.~Orbanz and D.~M. Roy, ``{Bayesian Models of Graphs, Arrays and Other
  Exchangeable Random Structures},'' {\em IEEE Transactions on Pattern Analysis
  and Machine Intelligence}, pp.~1--25, 2014.

\bibitem{Snijders2011}
T.~A.~B. Snijders, ``{Statistical Models for Social Networks},'' {\em Annual
  Review of Sociology}, vol.~37, no.~1, pp.~131--153, 2011.

\bibitem{asta2014geometric}
D.~Asta and C.~R. Shalizi, ``Geometric network comparison,'' {\em
  \textnormal{Preprint, }arXiv:1411.1350}, 2014.

\bibitem{Chatterjee2012}
S.~Chatterjee, ``{Matrix estimation by Universal Singular Value
  Thresholding},'' {\em Annals of Statistics, \textnormal{to appear},
  arxiv:1212.1247}.

\bibitem{Olhede2014a}
S.~C. Olhede and P.~J. Wolfe, ``{Network histograms and universality of
  blockmodel approximation},'' {\em PNAS}, vol.~111, p.~14722--14727, 2014.
  
\bibitem{Yang2014a}
J.~J. Yang, Q.~Han, and E.~M. Airoldi, ``{Nonparametric estimation and testing
  of exchangeable graph models},'' {\em AISTATS}, vol.~33, pp.~1060--1067,
  2014.

\bibitem{Bickel2011}
P.~J. Bickel, A.~Chen, and E.~Levina, ``{The method of moments and degree
  distributions for network models},'' {\em Annals of Statistics}, vol.~39,
  no.~5, pp.~2280--2301, 2011.

\bibitem{Wolfe2013}
P.~J. Wolfe and S.~C. Olhede, ``{Nonparametric graphon estimation},'' {\em
  \textnormal{Preprint, }arXiv: 1309.5936}, 2013.

\bibitem{Bickel2009}
P.~J. Bickel and A.~Chen, ``{A nonparametric view of network models and
  Newman--Girvan and other modularities},'' {\em Proc.\ Natl.\ Acad.\ Sci.\
  USA}, vol.~106, no.~50, pp.~21068--73, 2009.

\bibitem{Lloyd-randpriors-2012}
J.~R. Lloyd, P.~Orbanz, Z.~Ghahramani, and D.~M. Roy, ``{Random function priors
  for exchangeable arrays with applications to graphs and relational data},''
  {\em NIPS}, 2012.

\bibitem{Caron2014}
F.~Caron and E.~B. Fox, ``{Bayesian nonparametric models of sparse and
  exchangeable random graphs},'' in {\em NIPS Workshop on Frontiers in Network
  Analysis}, 2014.

\bibitem{Diaconis2007}
P.~Diaconis and S.~Janson, ``{Graph limits and exchangeable random graphs},''
  {\em Rendiconti di Matematica, Serie VII}, vol.~28, pp.~33--61, 2007.

\bibitem{Austin2008}
T.~Austin, ``{On exchangeable random variables and the statistics of large
  graphs and hypergraphs},'' {\em Probability Surveys}, vol.~5, pp.~80--145,
  2008.

\bibitem{Lovasz2012}
L.~Lov\'{a}sz, {\em {Large networks and graph limits}}.
\newblock American Mathematical Society, 2012.

\bibitem{Liben-Nowell2005}
D.~Liben-Nowell, J.~Novak, R.~Kumar, P.~Raghavan, and A.~Tomkins, ``{Geographic
  routing in social networks},'' {\em PNAS}, vol.~102, no.~33,
  pp.~11623--11628, 2005.

\bibitem{Eagle2010}
N.~Eagle, M.~Macy, and R.~Claxton, ``{Network diversity and economic
  development.},'' {\em Science}, vol.~328, no.~5981, pp.~1029--31, 2010.

\bibitem{Shalizi2011a}
C.~R. Shalizi and A.~C. Thomas, ``{Homophily and Contagion Are Generically
  Confounded in Observational Social Network Studies.},'' {\em Sociological
  methods \& research}, vol.~40, no.~2, pp.~211--239, 2011.

\bibitem{Fosdick2013}
B.~K. Fosdick and P.~D. Hoff, ``{Testing and Modeling Dependencies Between a
  Network and Nodal Attributes},'' {\em \textnormal{Preprint,
  }arXiv:1306.4708}, 2013.

\bibitem{VonLuxberg2012}
U.~von Luxberg, R.~C. Williamson, and I.~Guyon, ``{Clustering: Science or
  art?},'' {\em Journal of Machine Learning Research: W\&CP}, vol.~27,
  pp.~65--79, 2012.

\bibitem{Mccormick2013-ARD}
T.~H. McCormick and T.~Zheng, ``{Latent space models for networks using
  Aggregated Relational Data},'' tech. rep., University of Washington, 2013.

\bibitem{Yan2014}
X.~Yan, J.~E. Jensen, F.~Krzakala, C.~Moore, C.~R. Shalizi, L.~Zdeborov,
  P.~Zhang, and Y.~Zhu, ``{Model selection for degree-corrected block
  models},'' {\em J Stat Mech: Theory and Experiment}, pp.~1--13, 2014.

\bibitem{Peixoto2013}
T.~P. Peixoto, ``{Parsimonious module inference in large networks},'' {\em Phys
  Rev Lett}, p.~148701, 2013.

\bibitem{peixoto2014model}
T.~P. Peixoto, ``Model selection and hypothesis testing for large-scale network
  models with overlapping groups,'' {\em \textnormal{Preprint,
  }arXiv:1409.3059}, 2014.

\bibitem{Hofman2008}
J.~Hofman and C.~Wiggins, ``{Bayesian Approach to Network Modularity},'' {\em
  Phys Rev Lett}, vol.~100, no.~25, p.~258701, 2008.

\bibitem{Gopalan2012}
P.~Gopalan, D.~Mimno, S.~M. Gerrish, M.~J. Freedman, and D.~M. Blei,
  ``{Scalable inference of overlapping communities},'' {\em NIPS}, 2012.

\bibitem{Hunter2008}
D.~R. Hunter, S.~M. Goodreau, and M.~S. Handcock, ``{Goodness of Fit of Social
  Network Models},'' {\em Journal of the American Statistical Association},
  vol.~103, no.~1, pp.~248--258, 2008.

\bibitem{Gelman2013}
A.~Gelman and C.~R. Shalizi, ``{Philosophy and the practice of Bayesian
  statistics},'' {\em British Journal of Mathematical and Statistical
  Psychology}, vol.~66, no.~1996, p.~36, 2013.

\bibitem{Shalizi2013a}
C.~R. Shalizi and A.~Rinaldo, ``{Consistency under sampling of exponential
  random graph models},'' {\em Annals of Statistics}, vol.~41, no.~2,
  pp.~508--535, 2013.

\bibitem{Salter-Townshend2013}
M.~Salter-Townshend and T.~B. Murphy, ``{Variational Bayesian inference for the
  Latent Position Cluster Model for network data},'' {\em Computational
  Statistics \& Data Analysis}, vol.~57, no.~1, pp.~661--671, 2013.

\bibitem{Reed2013}
C.~Reed, {\em {Submodular MAP Inference for Scalable Latent Feature Models}}.
\newblock PhD thesis, University of Cambridge, 2013.

\bibitem{Amini2013}
A.~A. Amini, A.~Chen, P.~J. Bickel, and E.~Levina, ``{Pseudo-likelihood methods
  for community detection in large sparse networks},'' {\em Annals of
  Statistics}, vol.~41, no.~4, pp.~2097--2122, 2013.

\bibitem{Raftery2012}
A.~E. Raftery, X.~Niu, P.~D. Hoff, and K.~Y. Yeung, ``{Fast Inference for the
  Latent Space Network Model Using a Case-Control Approximate Likelihood},''
  {\em Journal of Computational and Graphical Statistics}, vol.~21, no.~4,
  pp.~901--919, 2012.

\bibitem{Nishihara2013}
R.~Nishihara, T.~Minka, and D.~Tarlow, ``{Detecting Parameter Symmetries in
  Probabilistic Models},'' {\em \textnormal{Preprint, }arXiv:1312.5386}, 2013.

\end{thebibliography}

}
\end{document}